\documentclass[sigconf]{acmart}
\AtBeginDocument{%
  }

\acmSubmissionID{2671}

\usepackage[dvipsnames]{xcolor}
\usepackage{array}
\usepackage{xspace}
\usepackage{multirow}
\usepackage{tcolorbox}
\tcbuselibrary{breakable}  
\newcommand{\gear}{\textsc{GeAR}\xspace}
\begin{document}

\title{Millions of \gear-s \includegraphics[height=\baselineskip]{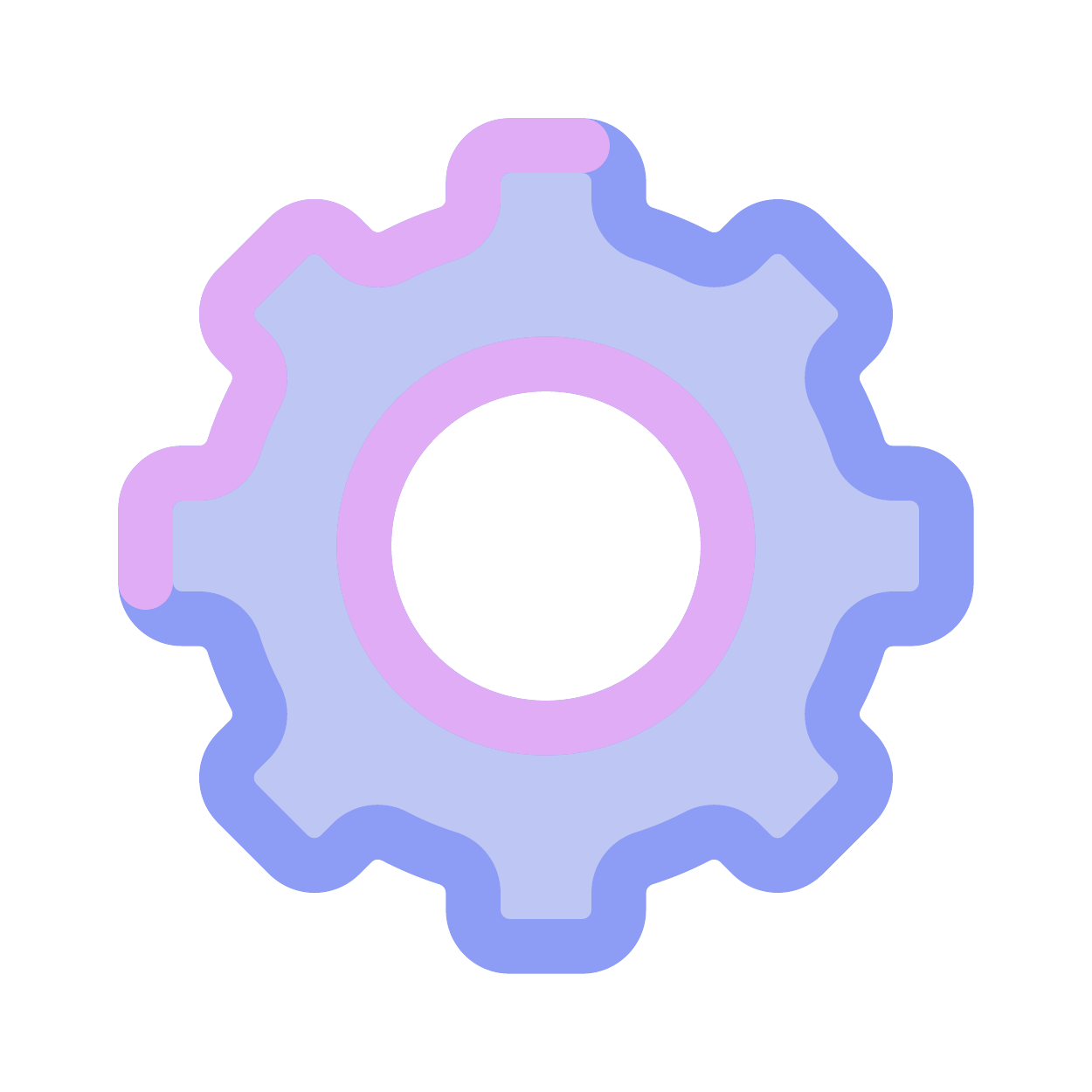}: Extending GraphRAG to Millions of Documents}

\newcolumntype{L}[1]{>{\raggedright\let\newline\\\arraybackslash\hspace{0pt}}m{#1}}
\newcolumntype{C}[1]{>{\centering\let\newline\\\arraybackslash\hspace{0pt}}m{#1}}
\newcolumntype{R}[1]{>{\raggedleft\let\newline\\\arraybackslash\hspace{0pt}}m{#1}}

\author{Zhili Shen}
\email{Zhili.Shen17@gmail.com}
\affiliation{%
  \institution{Huawei Technologies Co., Ltd.}
  \city{Edinburgh}
  \country{United Kingdom}
}
\author{Chenxin Diao}
\email{chenxindiao@huawei.com}
\affiliation{%
  \institution{Huawei Technologies Co., Ltd.}
  \city{Edinburgh}
  \country{United Kingdom}
}

\author{Pascual Merita}
\email{pascual.merita@h-partners.com}
\affiliation{%
  \institution{Huawei Technologies Co., Ltd.}
  \city{Edinburgh}
  \country{United Kingdom}
}

\author{Pavlos Vougiouklis}
\email{pavlos.vougiouklis@huawei.com}
\affiliation{%
  \institution{Huawei Technologies Co., Ltd.}
  \city{Edinburgh}
  \country{United Kingdom}
}

\author{Jeff Z. Pan}
\email{j.z.pan@ed.ac.uk}
\affiliation{%
  \institution{University of Edinburgh}
  \city{Edinburgh}
  \country{United Kingdom}
}

\newtcolorbox[list inside=prompt,auto counter,number within=section]{prompt}[1][]{
    colbacktitle=black!60,
    coltitle=white,
    fontupper=\footnotesize,
    boxsep=5pt,
    left=0pt,
    right=0pt,
    top=0pt,
    bottom=0pt,
    boxrule=1pt,
    title={#1},
    breakable,
    #1, 
}
\settopmatter{printacmref=true} 
\renewcommand\footnotetextcopyrightpermission[1]{} 

\renewcommand{\shortauthors}{Shen et al.}

\begin{abstract}
Recent studies have explored graph-based approaches to retrieval-augmented generation, leveraging structured or semi-structured information---such as entities and their relations extracted from documents---to enhance retrieval. However, these methods are typically designed to address specific tasks, such as multi-hop question answering and query-focused summarisation, and therefore, there is limited evidence of their general applicability across broader datasets. In this paper, we aim to adapt a state-of-the-art graph-based RAG solution: \gear and explore its performance and limitations on the SIGIR 2025 LiveRAG Challenge.
\end{abstract}




\received{23 May 2025}

\maketitle

\section{Introduction}

Retrieval-augmented Generation (RAG) has demonstrated significant improvements in the performance of Large Language Models (LLMs) on Question Answering (QA) tasks \cite{Lewis2020}. While RAG is effective for handling single-hop queries, multi-hop QA remains a more complex problem, as it necessitates compositional reasoning over multiple retrieved passages or documents.

Recent studies have explored graph-based approaches for RAG, leveraging information, such as entities and their relations extracted from documents, to enhance
retrieval performance \cite{Fang2024,Gutierrez2024,Li2024,Shen2024}.
These methods---commonly referred to as GraphRAG---have achieved state-of-the-art performance across many multi-hop QA datasets, such as MuSiQue, HotpotQA, and 2WikiMultihopQA~\cite{Gutierrez2024,Shen2024}. However, they are typically applied to smaller-scale document datasets containing up to hundreds of thousands of passages, and, therefore, there is limited evidence supporting their applicability to larger or more diverse datasets. We took the opportunity to explore how our own GraphRAG approach: \gear \cite{Shen2024}, could be adapted to scale to the requirements of the millions of passages included in the FineWeb-10BT of the SIGIR 2025 LiveRAG Challenge.

Recent GraphRAG approaches, including \gear rely on the alignment of an index of passages with an index of triples extracted from these passages~\cite{Fang2024,Gutierrez2024,Li2024,Shen2024}. These triples represent atomic facts within their source passages and are then organised into a graph by connecting those that share common entities.
In GraphRAG settings, triple extraction is usually performed using LLM-based triple extraction methodologies. These schema-free Knowledge Graph (KG) construction methodologies have exhibited significant improvements in general domains that depart from the conventional ClosedIE or OpenIE settings, which are too constrained and unconstrained respectively in terms of named entities and pre-defined relations~\cite{Gutierrez2024}. However, running an LLM over millions or billions of passages entails significant costs, which prohibit the widespread adoption of such methodologies on web-scale corpora.

In our submitted solution, we seek to sidestep this \textit{offline} triple extraction step entirely. We adapt the agentic operations within \gear to iteratively pseudo-align passages retrieved during a baseline retrieving step (e.g., BM25) with triples from an existing KG, such as Wikidata. We \textit{expand} these triples forming candidate reasoning chains, which we, subsequently, use to retrieve additional passages across more distant reasoning paths with respect to the original input question.

We align Wikidata triples with FineWeb passages using conventional retrieval strategies which, while simple, proved surprisingly effective in our experiments. Based on the preliminary, automatic evaluation results our submission: ``Graph-Enhanced RAG'' achieved \textit{correctness} and \textit{faithfulness} scores of $0.875714$ and $0.529335$ respectively. Below, we summarise our key observations from the challenge and outline open questions for future research:

\begin{itemize}
\item While state-of-the-art GraphRAG methods have demonstrated superior performance in multi-hop reasoning, they do not scale easily to corpora containing millions or billions of documents.
\item We propose a simple yet effective, online approach for aligning an index of passages with triples from Wikidata, using Falcon-3B-Instruct as a \textit{knowledge synchroniser}.
\item We identify limitations in the current framework, re-iterating the need for improved asymmetric semantic models capable of operating within a shared semantic space for both graph data and text.
\end{itemize}

\section{Preliminaries}

Let $\mathbf{C} = \left \{c_1, c_2, \ldots, c_C \right \}$ be an index of textual passages (i.e. we would refer to them as \textit{chunks} interchangeably) and $\mathbf{q}$ be a natural language question that is provided as an input to our system. Retrieving items from $\mathbf{C}$ relevant to $\mathbf{q}$ can be achieved by using a base retrieval function $h^k_{\text{base}}\left( \mathbf{q}, {\mathbf{C}}\right ) \subseteq \mathbf{C}$ for returning a ranked list of items from $\mathbf{C}$, in descending order according to a particular retrieval score.

In the context of this challenge, given an input query $\mathbf{q'}$, a dense retriever on top of $\mathbf{C}$: $h^k_{\text{dense}}\left( \mathbf{q'}, {\mathbf{C}}\right )$ can be implemented using the provided Pinecone index. Similarly, a sparse retrieval step: $h^k_{\text{sparse}}\left( \mathbf{q'}, {\mathbf{C}}\right )$ can be achieved using the provided OpenSearch instance. A baseline retrieval step can be also implemented as a \textit{hybrid} combination of passages coming from a dense and sparse retriever. Each hybrid retrieval search step returns the top-$k$ items from an index of interest by aggregating the results of dense and sparse retrieval using Reciprocal Rank Fusion (RRF)~\cite{Cormack2009}, as follows: 
\begin{align}
h^k_{\text{hybrid}}\left( \mathbf{q'}, {\mathbf{C}}\right ) = \text{RRF}\left(h^k_{\text{dense}}\left( \mathbf{q'}, {\mathbf{C}}\right ), h^k_{\text{sparse}}\left( \mathbf{q'}, {\mathbf{C}}\right )\right),
\end{align}
where $h^k_{\text{dense}}, h^k_{\text{sparse}} \subseteq \mathbf{C}$ are functions
returning sets of items from $\mathbf{C}$, in descending order according to $\text{score}_{\text{dense}}$ and $\text{score}_{\text{sparse}}$ respectively.


\begin{figure}[t]
  \centering
  \includegraphics[width=\linewidth]{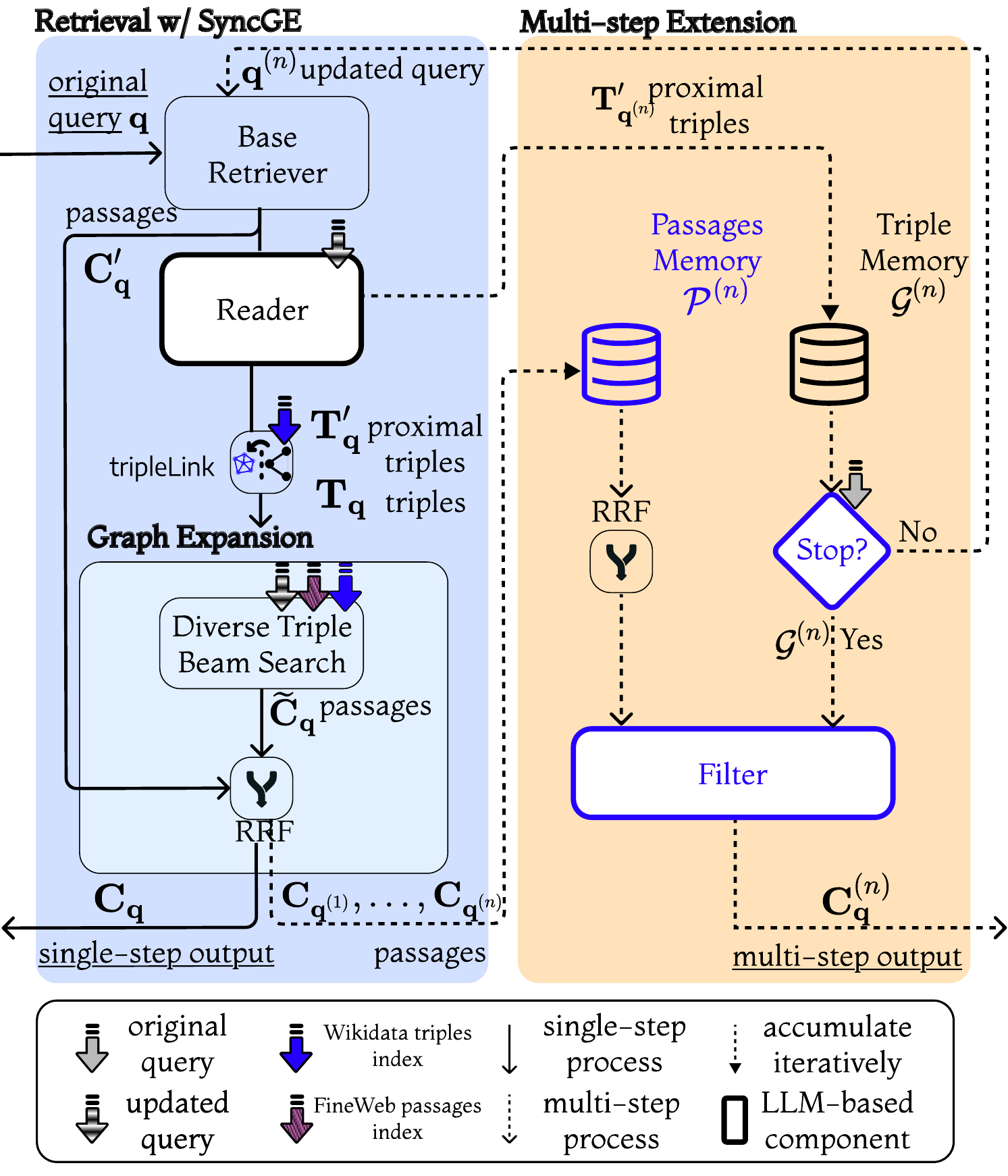}
  \caption{System Architecture. New or modified components in \gear are highlighted in \textcolor{blue}{blue}.}
  \label{fig:sys_diagram}
\end{figure}

\section{Extending \gear to Millions of Documents}
Our goal is to retrieve the most suitable passages from $\mathbf{C}$ that would enable a retrieval-augmented model that uses Falcon3-10B-Instruct as its reader to answer the input question using the context provided in the retrieved passages \cite{Lewis2020}. Our solution uses \gear as a backbone retriever, and we adapt it accordingly to the requirements of the task. As shown in Figure~\ref{fig:sys_diagram}, instead of relying on an explicit alignment between passages and triples, we propose a simple yet effective approach for pseudo-aligning passages retrieved during a baseline retrieving step with triples from Wikidata, without incorporating any offline association between the two indices. We use these triples to \textit{approximate} passages at more distant reasoning steps. 

To facilitate readability, we describe our solution, highlighting changes to \gear in \textcolor{blue}{blue}-coloured font. Some of these changes are due to architectural restrictions (i.e. the alignment of proximal triples from FineWeb with Wikidata in Eq.~\ref{eq:triple_alignment}) and some other because they were leading to improvements in our development set (see Section~\ref{subsec:using_datamorgana}).


\section{Multi-step Agentic Retrieval}
\label{sec:multi_step_approach}
By default \gear supports a multi-step, agentic framework that seeks to 
\begin{itemize}
\item reason over the cumulative collected evidence to determine termination
\item rewrite the query should additional retrieval steps be required for answering it successfully.
\end{itemize}

Within this multi-turn setting, the original input question $\mathbf{q}$ is iteratively decomposed into simpler queries: $\mathbf{q}^{(1)}, \ldots, \mathbf{q}^{(n)}$, where $\mathbf{q}^{(1)} = \mathbf{q}$ and $n \in \mathbb{N}$ represent the number of the current step. Following the \gear framework, at each step $n$, we retrieve a preliminary list of candidate passages: $\mathbf{C}_{\mathbf{q}^{(n)}}' = h^k_{\text{hybrid}}\left( \mathbf{q}^{(n)}, {\mathbf{C}}\right )$. We \textit{synchronise} this list of passages with the parametric LLM knowledge by returning a set of proximal triples that can support answering the current query (see Reader prompt at Appendix~\ref{subsec:gear_prompts}):
\begin{align}
    \mathbf{T}_{\mathbf{q}^{(n)}}' = \texttt{read}\left (\mathbf{C}_{\mathbf{q}^{(n)}}', {\mathbf{q}^{(n)}}\right ).
    \label{eq:proximal_read}
\end{align}
To facilitate the multi-step capabilities, our system maintains memory objects for retrieved passages and proximal triples: \textcolor{blue}{$\mathcal{P}^{(n)}$} and $\mathcal{G}^{(n)}$ respectively. $\mathcal{G}^{(n)}$ is updated after every \texttt{read} step (see Eq.~\ref{eq:proximal_read}) ensuring uniqueness of the enclosed triples.

\subsection{Leveraging an External Knowledge Graph}
Let $\mathbf{T} = \left \{t_1, t_2, \ldots,t_T: t_j = \left ( s_j, p_j, o_j \right ) \right \}$ be a knowledge graph of triples used in parallel with the provided FineWeb-
10BT chunks. \textcolor{blue}{In contrast to the original \gear setting, the triples in this challenge are not explicitly associated with the chunks in $\mathbf{C}$ nor directly extracted from them.} We link the proximal triples extracted above to triples in $\mathbf{T}$, as follows:
\begin{align}
    \mathbf{T}_{\mathbf{q}^{(j)}} =\left \{t_i | t_i = h^1_{\text{sparse}}\left( t_i', {\mathbf{T}}\right) ~ \forall t_i' \in \mathbf{T}_{\mathbf{q}^{(n)}}'\right \}.
    \label{eq:triple_alignment}
\end{align}
The triple linking mechanism can vary. However, in this paper, we consider it to be simply retrieving the most similar triple from $\mathbf{T}$ based on sparse similarity. We follow \gear to perform \textit{graph expansion} with diverse triple beam search (see Section 4.2 in \cite{Shen2024}) to return sequences of triples from $\mathbf{T}$ that are the most relevant to answering $\mathbf{q}^{(n)}$.

\begin{table*}[ht]
\footnotesize
\centering
\caption{Question and answer type taxonomy used by DataMorgana, along with their respective probability. Descriptions are summarised for conciseness.}
\label{tab:question_taxonomy}
\begin{tabular}{@{}llp{1cm}p{11.5cm}@{}}
\toprule
\textbf{Categorisation} & \textbf{Category} & \textbf{Prob.} & \textbf{Description} \\
\midrule
\multirow{10}{*}{\begin{tabular}[c]{@{}l@{}}Question\\Formulation\end{tabular}} 
& Concise and Natural & 10\% & A direct natural question consisting of a few words. \\
& Verbose and Natural & 10\% & A long question consisting of more than 9 words. \\
& List-based & 10\% & Asks for multiple items or examples. Often begins with `What are some' or `List the'. \\
& Definition-based & 10\% & Explicitly asks for meaning or definition of a term. Often begins with `What is' or `Define'. \\
& Opinion-seeking & 10\% & Asks for subjective viewpoints rather than facts. Includes phrases like `What do you think' or `Should we'. \\
& Hypothetical & 10\% & About imaginary scenarios. Often begins with `What if', `Imagine that', or `Suppose that'. \\
& How-to & 10\% & Seeks procedural knowledge or step-by-step guidance. Begins with `How to' or `How do I'. \\
& Yes/No & 10\% & Can be answered with 'yes' or 'no'. Often begins with `Is', `Are', `Do', `Can', `Will'. \\
\midrule
\multirow{2}{*}{\begin{tabular}[c]{@{}l@{}}Premise\\Categorisation\end{tabular}} 
& w/o Premise & 70\% & A question without any premise or information about the user. \\
& w/ Premise & 30\% & A question starting with a short premise revealing user needs or information. \\
\midrule
\multirow{5}{*}{\begin{tabular}[c]{@{}l@{}}Answer\\Type\end{tabular}} 
& Factoid & 15\% & Seeks specific, concise information like names, dates, or numbers about a particular subject. \\
& Multi-aspect & 25\% & About two different aspects of the same entity requiring information from two separate documents. \\
& Comparison & 30\% & Compares two related concepts by a common meaningful attribute using information from two documents. \\
& Path-following & 15\% & Requires following a clear, predefined reasoning path between entities to find the answer. \\
& Path-finding & 15\% & Requires identifying the correct path when many potential connections between entities exist. \\
\bottomrule
\end{tabular}
\end{table*}

However, since in contrast to the default \gear setting we do not have a direct alignment between the triples participating in the resulting beams with chunks in $\mathbf{C}$, we opt for a looser online alignment using a base retrieval strategy. After top beams are flattened in a
breadth-first order. \textcolor{blue}{Each triple in the resulting list is mapped to a chunk using $h^1_{\text{sparse}}\left( t_i', {\mathbf{C}}\right)$.} Let $\widetilde{\mathbf{C}}_{\mathbf{q}^{(n)}}$ be the list of unique passages after this soft alignment. The candidate list of passages at step $n$ is obtained using:
\begin{align}
    \mathbf{C}_{\mathbf{q}^{(n)}}^{\text{RRF}} = \mathrm{RRF}\left(\widetilde{\mathbf{C}}_{\mathbf{q}^{(n)}}, \mathbf{C}_{\mathbf{q}^{(n)}}'\right ).
\end{align}
The returned passages at this step are appended at the running passages memory $\mathcal{P}^{(n)}$.
\subsection{Query Re-writing and Termination}
The query re-writing process leverages Falcon3-10B-Instruct, and incorporates the triple memory and the entire query rewriting history up to the current $n$ step: $\mathcal{G}^{(n)}$ and $\mathbf{q}^{(1)}, \ldots, \mathbf{q}^{(n)}$ respectively. This process can be formally expressed as:
\begin{align}
\mathbf{d}^{(n)}, \mathbf{q}^{(n+1)} = \texttt{rewrite}\left (\mathcal{G}^{(n)}, \mathbf{q}^{(1)} \textcolor{blue}{, \ldots, \mathbf{q}^{(n)}} \right),
\label{eq:rewrite}
\end{align}
where $\mathbf{d}^{(n)} \in \left \{ \text{``Yes''}, \text{``No''} \right \}$ denotes the query's answerability and $\mathbf{q}^{(n+1)}$ represents the updated query, which serves as input for the retriever in the next iteration. When the query is deemed answerable, the system concludes its iterative process and $\mathbf{q}^{(n+1)} \in \emptyset$.
\subsection{After Termination}
\paragraph{Filtering Irrelevant Passages}Since we are expecting noisy outputs coming from this alignment strategy, we introduce a prompting stage that seeks to filter out irrelevant passages. \textcolor{blue}{The final list of returned passages that will be used for question answering, after termination, is formed as follows:
\begin{align}
    \mathbf{C}_{\mathbf{q}} = \texttt{filter}\left(\mathcal{P}^{(n)}, \mathcal{G}^{(n)}\right).
    \label{eq:filter}
\end{align}}\paragraph{Question Answering}In the final step, Falcon3-10B-Instruct is prompted to answer the original question $\mathbf{q}$ given $\mathbf{C}_{\mathbf{q}}$ and the accumulated triple memory $\mathcal{G}^{(n)}$ as follows:
\begin{align}
    \mathbf{a}_{\mathbf{q}} = \texttt{answer}\left(\mathbf{q}, \mathbf{C}_{\mathbf{q}}, \textcolor{blue}{\mathcal{G}^{(n)}}\right).
    \label{eq:question_answering}
\end{align}
All relevant prompts for the \texttt{read}, \texttt{rewrite}, \texttt{filter} and \texttt{answer} steps are provided in Appendix~\ref{sec:prompts}.

\section{Experiments}
For the external knowledge graph, we use the full Wikidata\footnote{\url{https://www.wikidata.org/wiki/Wikidata:Database_download}} dump, filtering out any triples whose object is a string literal. We include one alias\footnote{Including more aliases could improve performance, but we chose not to do so in order to stay within the provided compute credits and avoid incurring additional costs.} for each included entity by creating a separate triple with `alias' as predicate. We use a separate Pinecone sparse index for storing this data.

We set the maximum number of steps $n = 2$. Throughout our experiments, we identified that benefits of graph expansion for simpler questions coming from DataMorgana were limited. Consequently, we opted for a more efficient implementation that does not use Wikidata triples and graph expansion during the first iteration. Questions requiring multi-hop reasoning would require additional iterations, and consequently, the full pipeline described in Section~\ref{sec:multi_step_approach} is used for $n > 1$.

In order to monitor improvement in the pipeline, we built our evaluation according to the suggested methodology, focusing on correctness and faithfulness. Experiments were conducted by constructing a sample of questions using DataMorgana \cite{filice2025generatingdiverseqabenchmarks}.

\subsection{Using DataMorgana}
\label{subsec:using_datamorgana}

Following the best practices presented by \citeauthor{filice2025generatingdiverseqabenchmarks}, we divide the set of users into novice and expert with equal probability, and further define a set of question and answer type categories. We expand upon their original set by incorporating the `path-following' and `path-finding' multi-hop question categorisation introduced by \citeauthor{Gutierrez2024}. Moreover, we refrain from including the `linguistic variation' question type and redistribute their probability mass among the remaining categories. Table \ref{tab:question_taxonomy} presents our final taxonomy of question and answer types.

\section{Discussion}

\begin{table*}[htbp!]
\centering
\footnotesize
\caption{\label{tab:example_misalignment}Example of misalignment between FineWeb and Wikidata. The \textcolor{OliveGreen}{\underline{green}} keywords indicate the topic of the proximal triples, while the \textcolor{Mahogany}{\underline{red}} keywords indicate the topic of the linked Wikidata triples.}
\begin{tabular}{L{2.cm}L{7.2cm}|L{7.2cm}}
\toprule
\textbf{Question} & {Do frilled lizards and geoducks share any reproductive characteristics?} & {How come I always have to reset the high limit switch on my hot tub heater after draining and refillin the spa?}
 \\
\midrule
\textbf{Proximal triples $\mathbf{T}_{\mathbf{q}^{(n)}}'$ from \texttt{read}-ing FineWeb chunks} & [(\textcolor{OliveGreen}{\underline{`Pacific Geoducks'}}, `larvae swimming duration', `first 48 hours after hatching'), (\textcolor{OliveGreen}{\underline{`Pacific Geoducks'}}, `fertilization method', `external fertilization'), (\textcolor{OliveGreen}{\underline{`Pacific Geoducks'}}, `release eggs', `7 to 10 million eggs'), (\textcolor{OliveGreen}{\underline{`Pacific Geoducks'}}, `reproductive method', `broadcast spawning'), (\textcolor{OliveGreen}{\underline{`Pacific Geoducks'}}, `development stage', `develop a tiny foot and drop to the ocean floor in a few weeks')] & [(`faulty parts', `can cause', \textcolor{OliveGreen}{\underline{`high limit switch to trip'}}), (\textcolor{OliveGreen}{\underline{`high limit switch'}}, `trips due to', `water temperature exceeding safe limits'), (\textcolor{OliveGreen}{\underline{`primary operating thermostat'}}, `failure can lead to', `high limit switch tripping'), (\textcolor{OliveGreen}{\underline{`blocked or clogged vents'}}, `can cause', `high limit switch to trip'), (\textcolor{OliveGreen}{\underline{`thermistor'}}, `failure can lead to', `high limit switch tripping')]\\
\midrule
\textbf{Wikidata triples $\mathbf{T}_{\mathbf{q}^{(n)}}$ linked by proximal triples} &  [(`Larval development in the \textcolor{Mahogany}{\underline{Pacific oyster}} and the impacts of ocean acidification: Differential genetic effects in wild and domesticated stocks', `cites work', `Gene expression correlated with delay in shell formation in larval \textcolor{Mahogany}{\underline{Pacific oysters (Crassostrea gigas)}} exposed to experimental ocean acidification provides insights into shell formation mechanisms.'), (`\textcolor{Mahogany}{\underline{Egg consumption and risk of cardiovascular disease}}: three large prospective US cohort studies, systematic review, and updated meta-analysis', `cites work', `Land, irrigation water, greenhouse gas, and reactive nitrogen burdens of meat, eggs, and dairy production in the United States'), (`Cryptic diversity, geographical endemism and allopolyploidy in \textcolor{Mahogany}{\underline{NE Pacific seaweeds}}', `cites work', `Temporal windows of reproductive opportunity reinforce species barriers in a marine broadcast spawning assemblage.'), (`The Probable Method of Fertilization in Terrestrial \textcolor{Mahogany}{\underline{Hermit Crabs}} Based on a Comparative Study of Spermatophores', `published in', `Pacific Science'), (`Pacific', `located in the \textcolor{Mahogany}{\underline{administrative territorial entity}}', `Long Beach'), (`Genetic variation of wild and hatchery populations of the \textcolor{Mahogany}{\underline{Pacific oyster Crassostrea gigas}} assessed by microsatellite markers', `cites work', `Isolation and characterization of di- and tetranucleotide microsatellite loci in geoduck clams, Panopea abrupta.')] & 
[(`STUDY OF EFFECT OF CONSECUTIVE HEATING ON \textcolor{Mahogany}{\underline{THERMOLUMINESCENCE GLOW CURVES}} OF MULTI-ELEMENT TL DOSEMETER IN HOT GAS-BASED READER SYSTEM', `published in', `Radiation Protection Dosimetry'), (`Heat killing of \textcolor{Mahogany}{\underline{Bacillus subtilis spores}} in water is not due to oxidative damage', `cites work', `A superoxide dismutase mimic protects sodA sodB Escherichia coli against aerobic heating and stationary-phase death.'), (`Mineralogy of Sn–W–As–Pb–Zn–Cu-bearing alteration zones in intracontinental \textcolor{Mahogany}{\underline{rare metal granites (Central Mongolia)}}', `cites work', `The “chessboard” classification scheme of mineral deposits: Mineralogy and geology from aluminum to zirconium'), (`Water as a reservoir of \textcolor{Mahogany}{\underline{nosocomial pathogens}}', `cites work', `Superficial and systemic illness related to a hot tub'), (`Journal of Research of the U. S. Geological Survey, 1974, volume 2, issue 4', `has part(s)', `A \textcolor{Mahogany}{\underline{mineral separation procedure}} using hot Clerici solution'), (`Optimal \textcolor{Mahogany}{\underline{Water–Power Flow-Problem}}: Formulation and Distributed Optimal Solution', `published in', `IEEE transactions on control of network systems'), (`Hot tub-associated dermatitis due to \textcolor{Mahogany}{\underline{Pseudomonas aeruginosa}}. Case report and review of the literature', `published in', `Archives of Dermatology')]\\
\bottomrule
\end{tabular}
\end{table*}

To gain deeper insight into our system, we conducted a focused case study, as shown in Table~\ref{tab:example_misalignment}. Our analysis demonstrates that misalignment can arise when linking proximal FineWeb triples $\mathbf{T}_{\mathbf{q}^{(n)}}'$ to the corresponding Wikidata triples $\mathbf{T}_{\mathbf{q}^{(n)}}$. In both examples provided in Table~\ref{tab:example_misalignment}, the proximal triples identified during the \texttt{read} step are well aligned with the content of the FineWeb chunks. However, once linked to Wikidata, there is a clear divergence in topic. Specifically, in the first example, the topic shifts from `pacific geoducks' to `pacific oyster', while in the second example, it shifts from toiletry machinery to subjects related to geography and biology.

This issue is particularly significant, as it challenges a key assumption of the original \gear system---namely, that proximal triples can reliably serve as proxies for the ``real'' triples in the triple index. Our findings highlight the need for careful consideration in the linking process, as such misalignments may compromise the integrity and interpretability of the resulting knowledge graph. To this end, we believe that the sparse retrieval strategy employed in this submission serves as a strong baseline and highlights the need for more advanced semantic models capable of operating within a shared semantic space for both graph data and text.

\section{Conclusion}
We explore how a state-of-the-art GraphRAG method: \gear can be adapted to the context of datasets consisting of millions of passages. Our work is motivated by the fact that GraphRAG methodologies usually rely on LLM-based approaches for extracting triples from passages of interest---an approach that assumes highly capable LLMs, which are costly to run at scale.

We propose a simple yet effective online approach for aligning an index of passages with triples from Wikidata, and we identify limitations and failure cases in the current framework. Our findings underscore the need for improved asymmetric semantic models capable of operating within a shared semantic space for both graph data and text—an essential step toward extending the benefits of GraphRAG methods to large-scale tasks.

\bibliographystyle{ACM-Reference-Format}
\bibliography{bibliography}
\appendix
\clearpage
\onecolumn  
\section{Prompts}
\label{sec:prompts}
We use this section to list the prompts that were used for the ``Graph-Enhanced RAG'' submission to the SIGIR 2025 LiveRAG challenge.
\subsection{\gear Prompts}
 Similarly to the rest of the manuscript, any changes to the original \gear prompts are highlighted in \textcolor{blue}{blue} font.
\label{subsec:gear_prompts}

\begingroup\hypersetup{linkcolor=white}
\begin{prompt}[title={Reader\hfill{}(Eq.~\ref{eq:proximal_read}})]

Your task is to find unique facts that help answer an input question.\\
You should present these facts as knowledge triples, which are structured as (``subject'', ``predicate'', ``object'').\\

Example:\\
Question: When was Neville A. Stanton's employer founded?\\
Facts: (``Neville A. Stanton'', ``employer'', ``University of Southampton''), (``University of Southampton'', ``founded in'', ``1862'')\\

Now you are given some documents:\\
\texttt{\{retrieved\_docs\}}\\

Based on these documents find supporting unique fact(s) that may help answer the following question.\\
Note: if the information you are given is insufficient, output only the relevant unique facts you can find.\\

Question: \texttt{\{query\}}\\
Facts: 
\end{prompt}
\endgroup

\begingroup\hypersetup{linkcolor=white}
\begin{prompt}[title={Question Answering\hfill{}(Eq.~\ref{eq:question_answering}})]
As an advanced reading comprehension assistant, your task is to analyze text passages, knowledge triples, and corresponding questions meticulously, with the aim of providing the correct answer.\\
==================\\
For example:\\
==================\\
Wikipedia Title: Edward L. Cahn\\
Edward L. Cahn (February 12, 1899 – August 25, 1963) was an American film director.\\

Wikipedia Title: Laughter in Hell\\
Laughter in Hell is a 1933 American Pre-Code drama film directed by Edward L. Cahn and starring Pat O'Brien. The film's title was typical of the sensationalistic titles of many Pre-Code films. Adapted from the 1932 novel of the same name buy Jim Tully, the film was inspired in part by ``I Am a Fugitive from a Chain Gang'' and was part of a series of films depicting men in chain gangs following the success of that film. O'Brien plays a railroad engineer who kills his wife and her lover in a jealous rage and is sent to prison. The movie received a mixed review in ``The New York Times'' upon its release. Although long considered lost, the film was recently preserved and was screened at the American Cinematheque in Hollywood, CA in October 2012. The dead man's brother ends up being the warden of the prison and subjects O'Brien's character to significant abuse. O'Brien and several other characters revolt, killing the warden and escaping from the prison. The film drew controversy for its lynching scene where several black men were hanged. Contrary to reports, only blacks were hung in this scene, though the actual executions occurred off-camera (we see instead reaction shots of the guards and other prisoners). The "New Age" (an African American weekly newspaper) film critic praised the scene for being courageous enough to depict the atrocities that were occurring in some southern states.\\

Wikipedia Title: Theodred II (Bishop of Elmham)\\
Theodred II was a medieval Bishop of Elmham. The date of Theodred's consecration unknown, but the date of his death was sometime between 995 and 997.\\

Wikipedia Title: Etan Boritzer\\
Etan Boritzer (born 1950) is an American writer of children's literature who is best known for his book ``What is God?'' first published in 1989. His best selling ``What is?'' illustrated children's book series on character education and difficult subjects for children is a popular teaching guide for parents, teachers and child- life professionals. Boritzer gained national critical acclaim after ``What is God?'' was published in 1989 although the book has caused controversy from religious fundamentalists for its universalist views. The other current books in the ``What is?'' series include ``What is Love?'', ``What is Death?'', ``What is Beautiful?'', ``What is Funny?'', ``What is Right?'', ``What is Peace?'', ``What is Money?'', ``What is Dreaming?'', ``What is a Friend?'', ``What is True?'', ``What is a Family?'', ``What is a Feeling?''. The series is now also translated into 15 languages. Boritzer was first published in 1963 at the age of 13 when he wrote an essay in his English class at Wade Junior High School in the Bronx, New York on the assassination of John F. Kennedy. His essay was included in a special anthology by New York City public school children compiled and published by the New York City Department of Education.\\

Wikipedia Title: Peter Levin\\
Peter Levin is an American director of film, television and theatre.\\

\textcolor{blue}{Knowledge Triples:}\\
\textcolor{blue}{(Edward L. Cahn, born on, February 12, 1899)}\\
\textcolor{blue}{(Edward L. Cahn, profession, film director)}\\
\textcolor{blue}{(Edward L. Cahn, died on, August 25, 1963)}\\
\textcolor{blue}{(Edward L. Cahn, directed, Laughter in Hell)}\\
\textcolor{blue}{(Laughter in Hell, directed by, Edward L. Cahn)}\\
\textcolor{blue}{(Laughter in Hell, released in, 1933)}\\

Question: When did the director of film Laughter In Hell die?\\
Answer: The director of film Laughter In Hell, Edward L. Cahn, died on August 25, 1963.\\

==================
Now your turn.
==================

\end{prompt}
\endgroup

\subsection{Extra Prompts}
\label{subsec:extra_prompts}

\begingroup\hypersetup{linkcolor=white}
\begin{prompt}[title={Query Re-writing and Termination\hfill{}(Eq.~\ref{eq:rewrite}})]

Given a question and its associated retrieved knowledge triples, you are asked to evaluate if the triples by themselves are sufficient to formulate an answer to the original question (\{Yes\} or \{No\}).\\
Your answer must begin with \{Yes\} or \{No\}.\\

If \{Yes\}, just provide \{Yes\} without any additional content.

If \{No\}, please think about the additional evidence that needs to be found to answer the original question, and then provide a suitable next question for retrieving this potential evidence.\\
Note that you have access to all the question rewriting steps that have been performed already, if any.\\
Please make sure that the next question is different from all the previous questions. Break it down into smaller questions if needed.\\
As the number of question rewriting steps that have been performed already increases, the next question should be more vague, optimising for retrieving at least some evidence that is relevant to the original question.\\
Note that the next question must be included in separate curly brackets \{xxx\}.\\

Here are some examples:\\

\# Example 1:\\
Original Question: The Sentinelese language is the language of people of one of which islands in the Bay of Bengal?\\

Knowledge triples:\\
(Sentinelese language, Indigenous to, Sentinelese people)\\
(Bay of Bengal, area, Andaman and Nicobar Islands)\\

\# Answer:\\
\{Yes\}\\

\# Example 2:\\
Original Question: Who is the coach of the team owned by David Beckham?\\

Knowledge triples:\\
(David Beckham, co-owned, Inter Miami CF)\\
(David Beckham, country of citizenship, United Kingdom)\\

\# Answer:\\
\{No\} \{Who is the coach of Inter Miami CF?\}\\

Example 3:\\
Original Question: I read somewhere that the Civil War affected cotton trade. How much of England's cotton came from the US before the war?\\
Rewrites:\\
Rewrite 1: What percentage, or quantity, of England's cotton came from the US before the US Civil War? We need to focus on cotton trade to England at that time\\
Rewrite 2: England's cotton from the US before the US Civil War\\

Knowledge triples:\\
(American Civil War, resulted in, expansion of cotton production)\\
(Egypt, regarded as the best alternative, Egyptian cotton)\\
(British companies, began investing heavily in, cotton production in Egypt)\\

\# Answer:\\
\{No\} \{England's cotton imports\}\\

Now, please carefully consider the following case:\\

Question History:\\
Original Question: \texttt{\{query\}}\\
\texttt{\{query\_rewriting\_history\}}\\

Knowledge triples:\\
\texttt{\{triples\}}\\

\# Answer: 
\end{prompt}
\endgroup

\begingroup\hypersetup{linkcolor=white}
\begin{prompt}[title={Re-rank and Filter\hfill{}(Eq.~\ref{eq:filter}})]

You are an advanced assistant that can rank passages based on their relevance to the query.\\
Each passage is indicated by a number identifier []. Please rank them based on their relevance to query.\\
Please return the passages in descending relevance order using identifiers, where the most relevant passages should be listed first, and the output format is [] > [] > etc, e.g., [4] > [6] > etc.\\
If any passages are irrelevant, please remove their identifier completely from results. Return `None` if there are no relevant passages.\\

We also give you a list of knowledge triples which be think are relevant to the query. Use them to help you rank the passages.\\

==================\\
For example:\\
==================\\
Question: When did the director of film Laughter In Hell die?\\

Knowledge Triples:\\

(Edward L. Cahn, born on, February 12, 1899)\\
(Edward L. Cahn, profession, film director)\\
(Edward L. Cahn, died on, August 25, 1963)\\
(Edward L. Cahn, directed, Laughter in Hell)\\
(Laughter in Hell, directed by, Edward L. Cahn)\\
(Laughter in Hell, released in, 1933)\\

Passages:\\

[1] Wikipedia Title: Laughter in Hell\\
Laughter in Hell is a 1933 American Pre-Code drama film directed by Edward L. Cahn and starring Pat O'Brien. The film's title was typical of the sensationalistic titles of many Pre-Code films. Adapted from the 1932 novel of the same name buy Jim Tully, the film was inspired in part by ``I Am a Fugitive from a Chain Gang'' and was part of a series of films depicting men in chain gangs following the success of that film. O'Brien plays a railroad engineer who kills his wife and her lover in a jealous rage and is sent to prison. The movie received a mixed review in ``The New York Times'' upon its release. Although long considered lost, the film was recently preserved and was screened at the American Cinematheque in Hollywood, CA in October 2012. The dead man's brother ends up being the warden of the prison and subjects O'Brien's character to significant abuse. O'Brien and several other characters revolt, killing the warden and escaping from the prison. The film drew controversy for its lynching scene where several black men were hanged. Contrary to reports, only blacks were hung in this scene, though the actual executions occurred off-camera (we see instead reaction shots of the guards and other prisoners). The ``New Age'' (an African American weekly newspaper) film critic praised the scene for being courageous enough to depict the atrocities that were occurring in some southern states.\\

[2] Wikipedia Title: Theodred II (Bishop of Elmham)\\
Theodred II was a medieval Bishop of Elmham. The date of Theodred's consecration unknown, but the date of his death was sometime between 995 and 997.\\

[3] Wikipedia Title: Edward L. Cahn\\
Edward L. Cahn (February 12, 1899 – August 25, 1963) was an American film director.\\

[4] Wikipedia Title: Etan Boritzer\\
Etan Boritzer (born 1950) is an American writer of children's literature who is best known for his book ``What is God?'' first published in 1989. His best selling ``What is?'' illustrated children's book series on character education and difficult subjects for children is a popular teaching guide for parents, teachers and child- life professionals. Boritzer gained national critical acclaim after ``What is God?'' was published in 1989 although the book has caused controversy from religious fundamentalists for its universalist views. The other current books in the ``What is?'' series include ``What is Love?'', ``What is Death?'', ``What is Beautiful?'', ``What is Funny?'', ``What is Right?'', ``What is Peace?'', ``What is Money?'', ``What is Dreaming?'', ``What is a Friend?'', ``What is True?'', ``What is a Family?'', ``What is a Feeling?''. The series is now also translated into 15 languages. Boritzer was first published in 1963 at the age of 13 when he wrote an essay in his English class at Wade Junior High School in the Bronx, New York on the assassination of John F. Kennedy. His essay was included in a special anthology by New York City public school children compiled and published by the New York City Department of Education.\\

[5] Wikipedia Title: Peter Levin\\
Peter Levin is an American director of film, television and theatre.\\

Reranked Passages: [3] > [1]\\
==================\\

The following are \texttt{\{num\_docs\}} passages, each indicated by number identifier [].\\
Please rank them based on their relevance to query.\\
Please return the passages in descending relevance order using identifiers, where the most relevant passages should be listed first, and the output format is [] > [] > etc, e.g., [4] > [6] > etc.\\
If any passages are irrelevant, please remove their identifier completely from results. Return `None` if there are no relevant passages.\\

Question: \texttt{\{query\}}\\

Knowledge Triples:\\

\texttt{\{triples\}}\\

Passages:\\

\texttt{\{retrieved\_docs\}}\\

Reranked Passages:
\end{prompt}
\endgroup
\end{document}